\documentclass{article}
\usepackage{amsmath,epsfig}
\usepackage[preprint]{spconfa4}

\makeatletter

\newcommand{\Rmnum}[1]{\expandafter\@slowromancap\romannumeral #1@}
\makeatother

\usepackage{float}

\usepackage{xcolor}

\let\OLDthebibliography\thebibliography
\renewcommand\thebibliography[1]{
  \OLDthebibliography{#1}
  \setlength{\parskip}{0pt}
  \setlength{\itemsep}{0pt plus 0.3ex}
}

\begin{document}\sloppy

\def\x{{\mathbf x}}
\def\L{{\cal L}}

\title{Medium Transmission Map Matters for Learning to \\Restore Real-world Underwater Images}
%
\name{Kai Yan$^*$, Lanyue Liang$^*$,Ziqiang Zheng†, Guoqing Wang†, Yang Yang‡}
\address{$^*$kyan8@uic.edu; $^*$moonyue7070@gmail.com; †zhengziqiang1@gmail.com; \\
†gqwang0420@hotmail.com; ‡ dlyyang@gmail.com}

\maketitle

\begin{abstract}
Underwater visual perception is essentially important for underwater exploration, archeology, ecosystem and so on. The low illumination, light reflections, scattering, absorption and suspended particles inevitably lead to the critically degraded underwater image quality, which causes great challenges on recognizing the objects from the underwater images. The existing underwater enhancement methods that aim to promote the underwater visibility, heavily suffer from the poor image restoration performance and generalization ability. To reduce the difficulty of underwater image enhancement, we introduce the media transmission map as guidance to assist in image enhancement. We formulate the interaction between the underwater visual images and the transmission map to obtain better enhancement results. Even with simple and lightweight network configuration, the proposed method can achieve advanced results of 22.6 dB on the challenging Test-R90 with an impressive 30 times faster than the existing models. Comprehensive experimental results have demonstrated the superiority and potential on underwater perception. Paper's code is offered on: https://github.com/GroupG-yk/MTUR-Net.

\end{abstract}
\section{Introduction}
\label{sec:intro}
 With the development of science and technology, underwater research activities are also increasing, such as underwater object detection and tracking ~\cite{lee2012vision}, underwater robots~\cite{UDCP}and underwater monitoring~\cite{mohamed2011sensor}. However, the light reflections, scattering, absorption and suspended particles inevitably result in poor visibility with inhomogeneous illumination in the collected underwater images. In detail, the light is absorbed and scattered by suspended particles in the underwater setting, resulting in hazy effects on the images captured by the cameras. Water also attenuates light as a function of its salinity, light wavelength and depth since the red light is more attenuated due to a longer wavelength. Besides, the light intensity decreases with the increase of water depth. Such properties reduce visibility underwater and hamper the applicability of computer vision methods.

\begin{figure}[t]
    \centering
    \includegraphics[width=8cm]{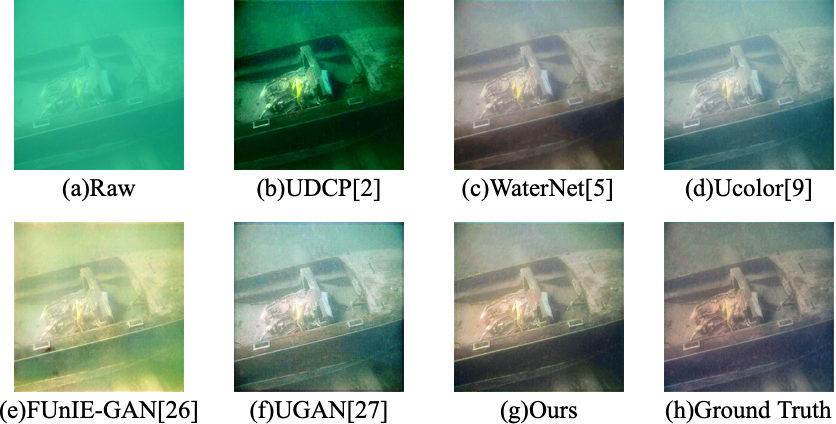}
    \caption{Comparison of the results of different methods for processing a real underwater picture. It can be seen from the results that our method restores the chromatic aberration and enhances the contrast.}
    \label{fig:survey}
\end{figure}

Early single-image underwater image restoration work used traditional physical methods to directly change the pixel value of the image~\cite{UDCP}~\cite{Histogram}. However, these methods have limited capabilities when faced with diverse underwater environments. Recently, driven by the release of a series of paired training sets including~\cite{water-net}~\cite{EUVP}~\cite{SQUID}, deep convolution neural networks (CNN) based models have been proposed by learning the mapping between underwater images and restored images. Representative methods include the WaterGAN Li et al.\cite{WaterGAN} and Ucolor by Li et al.\cite{Ucolor}, which consider the restoration in multi-channel spaces, and better results are obtained when compared the traditional physical designs as in Fig.\ref{fig:survey}. However, the quality improvement is limited due to the ignorance of other factors, such as distance-dependent attenuation and scattering. Considering the underwater imaging process, these factors can be considered by utilizing the semantics contained in the medium transmission map\cite{Ucolor}, such as the design proposed in this paper. By analyzing the results in Fig.\ref{fig:survey}(e), the improvement by the medium transmission map can be fully reflected by producing more visually pleasing results in terms of color, contrast, and naturalness.

In this work, our goal is to eliminate the influence of light scattering and attenuation on underwater images in real time to support intelligent underwater perception systems. Inspired by the depth-guided deraining model by Hu et al.~\cite{8953954}, we introduce the medium transmission map (MT) and formulate a MT-guided restoration framework. Specifically, a multi-task learning network is designed to generate both the MT and restoration outputs jointly. A multi-level (including both feature level and output level) knowledge interaction mechanism is proposed for better mining the guidance from the MT learning space. Furthermore, to maximally reduce the computational burden caused by the MT learning branch, parameters in some specific stages are shared across these two related tasks, thus enabling a real-time process of the underwater images.

In summary, this work has the following contributions:

\begin{itemize}
  \item We re-examined how to better use the medium transmission map. We can get good results by relying on RGB map alone using various preprocessing and color embedding, proving that MT map is of great significance for learning a more powerful real-world underwater image restoration network.
  \item A multi-task learning framework is formulated for leveraging the MT map, and a novel multi-level knowledge interaction mechanism is proposed for better mining the guidance from the MT learning space. 
  \item Comparative study on two real-world benchmarks demonstrated the superiority of our MTUR-Net over the state-of-the-art in terms of both restoration quality and inference speed.
\end{itemize}

The rest of our paper is organized as follows. Section~\ref{sec:related} briefly introduces the existing underwater image enhancement methods. Section~\ref{sec:method} presents the proposed underwater enhancement algorithm. The experimental results are reported in Section~\ref{sec:exper}, followed by the conclusion in Section~\ref{sec:conclusion}.

\section{Related Work}\label{sec:related}
\begin{figure*}
\centering
\includegraphics[width=16cm]{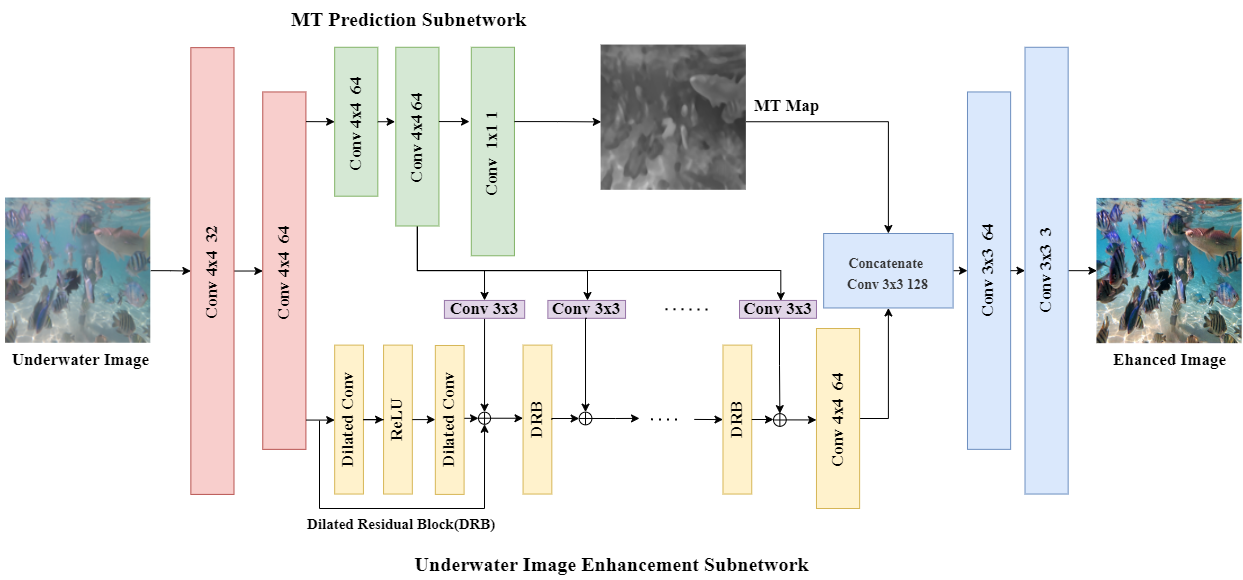}
\caption{Schematic diagram of  MTUR-Net. It consists of an encoder-decoder network for predicting MT map (green), a set of dilated residual blocks (yellow) to generate local features, convolutional layer (purple) for process MT features before fusion, and the convolutional layer (blue part) to upsample the feature map and generate underwater enhanced images. $\oplus$ pixel-wise addition}
\label{fig:picture001}
\end{figure*}


\textbf{Physical prior based methods}. Based on adjusting the pixel value to improve visual quality originally, and physical model-based methods are used widely before long, which have obtained impressive results, while still exist some shortcomings, that they are almost slow work and sensitive to different kinds of underwater images.

Recently, the development of scientific and technological artificial intelligence, the method based on deep learning has achieved remarkable results. Underwater image enhancement framework is mostly based on convolutional neural network(CNN) or generative adversarial network(GAN). For example, Li \textit{et al.}~\cite{water-net} proposed a simple CNN mdoel named Water-Net using gated fusion. Li et al.\cite{UWCNN} proposed UWCNN that based on underwater scene prior. Li et al. \cite{Ucolor} proposed an underwater image enhancement network: embedding a multi-color space via medium transmission-guided. 

J. Li et al.\cite{WaterGAN} used GANs and image formation models for supervised learning. To avoid requiring paired training data, it was proposed that a weakly supervised underwater color correction network (UCycleGAN) in \cite{UCycleGAN}. A multiscale dense GAN for powerful underwater image enhancement was described in \cite{MDGAN}.

Including the above research, these underwater image enhancement models often overlook the most important point, which focus in the real underwater environment, serving data under real conditions. For instance. \cite{UCycleGAN} use CycleGAN \cite{CycleGAN} network structure directly, and a simple multi-scale convolutional network is used in \cite{water-net}. in \cite{UWCNN} , faced  an underwater image of input, how to select the corresponding UWCNN model is challenging. \cite{Ucolor} is still not absolutely effective under the real underwater conditions.

In contrast to the above, our method has the following characteristic$\colon$(1) We trained and learned deep-guided non-local features and regressed the residual mapping to produce a clear output image. (2) our method adopts end-to-end training and is adaptable and convenient for most underwater scene. and, (3) our method achieves perfect performance on real underwater image datasets which is better than recently state-of-the-art methods.
\section{METHODOLOGY}~\label{sec:method}
Fig.~\ref{fig:picture001} shows the overall architecture of our medium transmission map guided underwater image restoration network (MTUR-Net). This network takes underwater images as input, and predicts the corresponding MT map and underwater enhanced images as output in an end-to-end manner. In general, the network first uses CNN to extract semantics and generate feature maps and share weights. Then two decoding branches are generated. (i) The MT prediction subnet, which uses the encoding and decoding network, to regress a medium transmission map from the input. (ii) The underwater image enhancement network, guided by the predicted MT map, predicts the enhanced image from the input underwater image.

\subsection{MT Prediction Subnet}
We review the haze removal method based on dark Channel prior\cite{he2010single}, which is widely used in harsh visual scenarios such as fog, dust and underwater.\cite{huang2014visibility}\cite{yang2011low}\cite{zhao2015deriving}.The image formation model can be expressed as\cite{fattal2008single}:

\begin{equation} I^{c}(x) = J^{c}(x)T(x) + A^{c}\big (1-T(x)\big),\quad c \in \{r, g, b\} \end{equation}

This equation is defined on three RGB color channels. $I$ represents the observed image, $A $ is the airlight color vector, $J$ is the surface brightness vector at the intersection of the scene and the real world light corresponding to the pixel x = (x,y), and $T(x)$ is the transmission along the light. And Y. -T. Peng et al. \cite{8307410} proposed a new Dark Channel Prior (DCP) algorithm that can effectively estimate ambient light and is suitable for enhancing foggy, hazy, sandstorm, and underwater images. Inspired by DCP, transferred T (X) has wide applicability, we use the medium transmission (MT) map ($\overline{T}$) as our attention map. It's effectiveness will be demonstrated in ablation experiments. From\cite{8307410}, the actual input underwater image does not  have a corresponding ground true medium transmission map,  it is difficult to train a deep neural network to estimate the  medium transmission map. So the medium transmission map can be estimated as:


\begin{equation} \widetilde {T}(x) =1-\min _{y \in \Omega (x)} \left \lbrace{ \min _{c \in \{r, g, b\}} \frac {I^{c}(y)}{A^{c}} }\right \rbrace \!. \end{equation}

$\tilde{T}$ is the estimated medium transmission map, $\Omega(x)$ is a local patch  centered at x and c is RGB channel.The schematicdiagram of the proposed module using the MT map is shown in Fig. \ref{fig:f1}. We use MT map as a feature selector to weigh the importance of different spatial locations of features, as shown in Fig. \ref{fig:f1}. Assign more weight to high-quality pixels (pixels  larger MT values), which can be expressed as:
\begin{figure}[h]
    \centering
    \includegraphics[width=6cm]{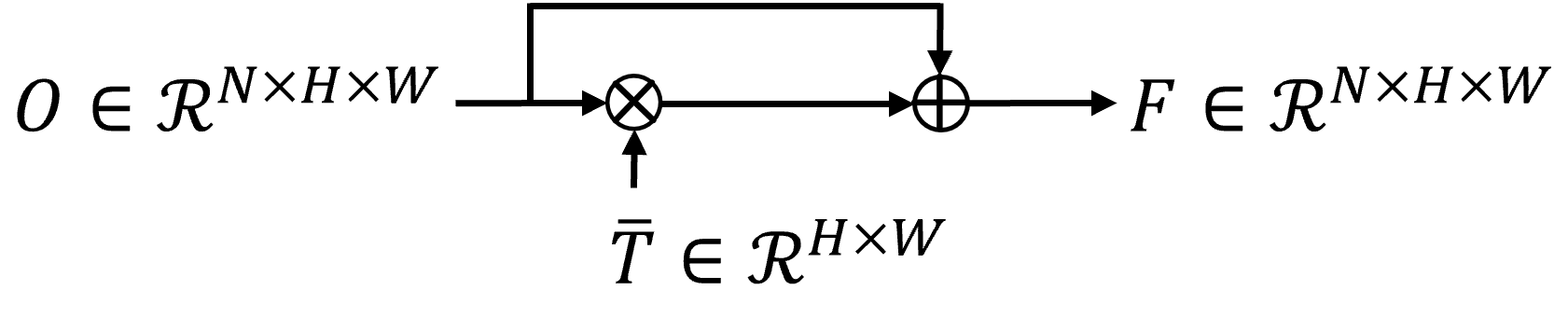}
    \caption{Medium transmission guidance module.The MT map $\overline{T}$ is a feature selector, $\overline{T}$ weighs the importance of the different spatial positions for $F$. }
    \label{fig:f1}
\end{figure}

\begin{equation}  {F} = {O}\oplus  {O}\otimes \overline {T}\end{equation}

$F,O$  represent the characteristics of the output and input respectively.
In detail, the MT map prediction sub-network uses 4 blocks to extract features. Each block has a convolution operation, a group normalization \cite{wu2018group} and a proportional exponent linear unit (SELU)\cite{SELU}
Then, it uses lateral connections to influence the detailed information decoded in the underwater feature map. Finally, another convolution operation is used, plus a sigmoid function, to return to $\overline{T}$ by adding a supervision (input MT map in the training data).

\subsection{Underwater Image Enhancement Subnet}
In the underwater image enhancement subnet, we use the convolution to reduce the resolution of the feature map, Then, followed by 11 dilated residual blocks(DRB) \cite{yu2017} to Increase the size of the perceptual field out reducing the resolution. Each DRB has a 3 * 3 dilated convolution \cite{chen2017deeplab}, a ReLU nonlinear function, and another 3 * 3  dilated convolution that adds input and output feature maps using skip connections. To avoid gliding issues, we set the dilation ratio of these 11 DRBs as {1,1,2,2,4,8,4,2,2,1} according to\cite{wang2018understanding}. Moreover, use the horizontally connected convolution module to add the MT prediction feature to the output feature map. After that, we use convolution to change the feature map to the size of the MT map and concatenate them together. Finally, through the convolution operation, scale the feature map to the size of the input image.

\begin{figure*}[t]
\centering
\includegraphics[width=16cm]{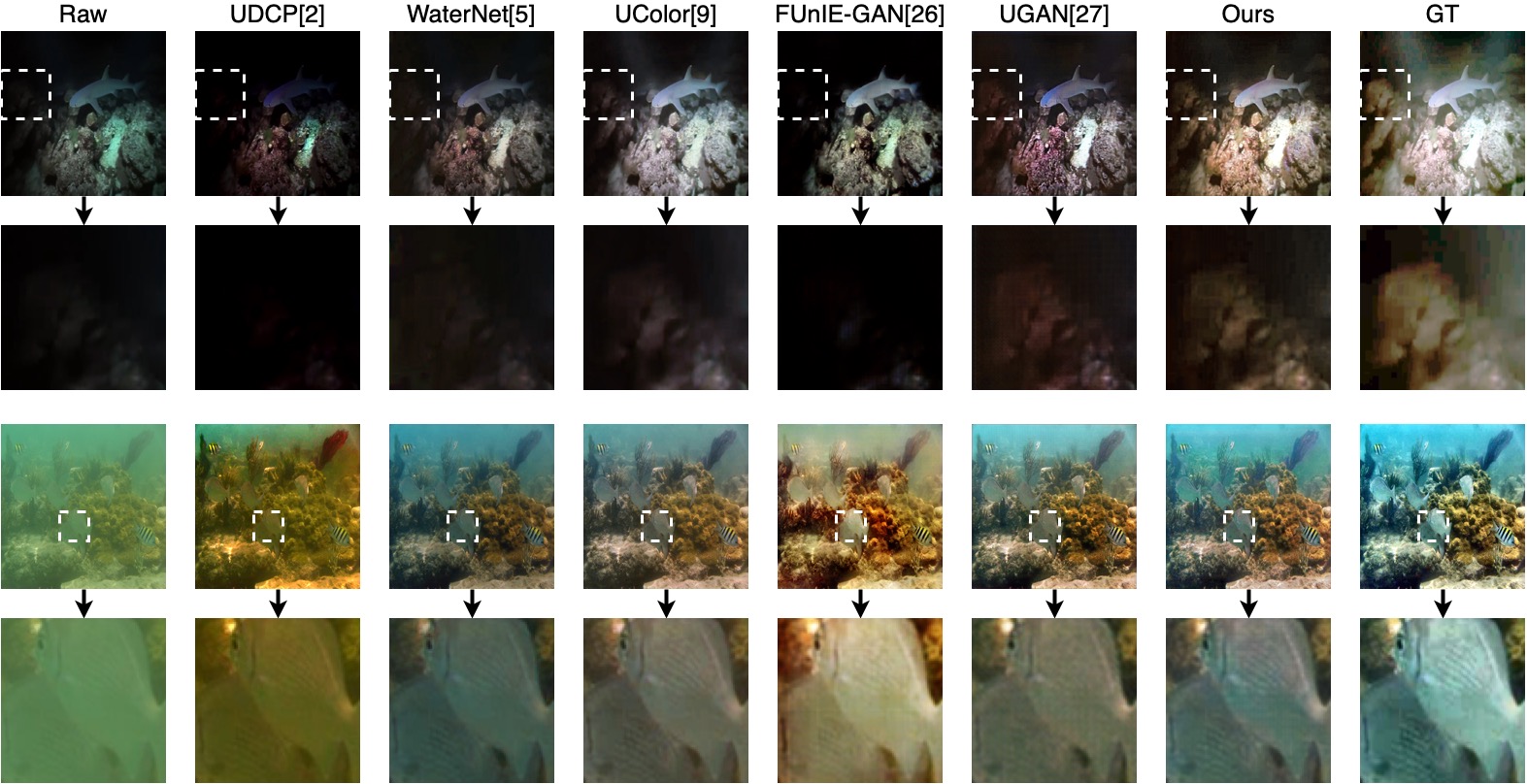}
\caption{Visual comparison of different images (from Test-R90) enhanced by state-of-the-art methods and our MTUR-Net.}
\label{fig:picture002}
\end{figure*}

\section{EXPERIMENTS}
In this section, we will first illustrate the details of the parameter design and then explain the settings of the entire experimental process. Above all, we compare our model with several existing models that performed well and provide ablation experiments at the end of this section to study the effective parts of MTUR-Net.

\subsection{Parameter Settings}
To train the network, we chose the real underwater image dataset illuminated in Li et al. \cite{Ucolor}, which contains 890 pairs of images from \cite{water-net}  and 1250 pairs of images from \cite{UWCNN}. We trained our network on a single NVIDIA 3090 Ti GPU with a batch size of 8, the initial learning rate is set to 1e-3, and network optimization is carried out by Adam.

\subsection{Experiment Setup}\label{sec:exper}
To test the proposed model, we took the remaining 90 pairs of real data in UIEB and recorded them as Test-R90, and to synthesize the multi-faceted results, we also tested 60 challenging images in UIEB, which were recorded as Test-C60.

To prove the advancement of this proposed model, we compared our method with other SOTA, including a physical model-based model and Deep-learning-based model. For the physical model is an extension of their previous work to deal with underwater image restoration called Underwater Dark Channel Prior(UDCP)\cite{UDCP}. What's more, Water-Net\cite{water-net}, a simple CNN model through gated fusion, Ucolor\cite{Ucolor}, a network embedding with the color space guided by media transmission, while a fully-convolutional conditional GAN-based model FUnIE-GAN\cite{FUnIE-GAN}, and a method using Generative Adversarial Networks (GANs) we chose\cite{UGAN}. To control variables, we chose the same training data and loss function as MTUR-Net.

\subsection{Comparitive Study}
In this experiment, we choose two evaluation methods, including the visual evaluation and quantitative evaluation, to compare the specific effects of our model with other models.

\textbf{Visual Evaluation}. 
In open water, due to the longest wave tension and fast propagation speed, red light compared to other wavelengths is absorbed more. Therefore, the underwater image appears blue or green. In order to clearly observe the effect of the image via  MTUR-Net processing, we provide a comparison chart of the corresponding results obtained in different ways. Fig. \ref{fig:picture002} shows that the output obtained by MTUR-Net has the best performance.Our solution can repair the chromatic aberration caused by different water areas and see the details in the dark water and the texture of fish in the muddy water on the restored image.

\textbf{Quantitative Evaluation}.
We provide full-reference evaluation and non-reference evaluation to quantitatively analyze the performance of different methods.

\begin{table}

\begin{center}

\caption{Comparison  the State-of-the-Arts Using the PSNR and SSIM on the Test-R90 Dataset~\cite{water-net}}  \vspace{10pt}\label{tab:r90}
\resizebox{\linewidth}{!}{
\begin{tabular}{c|c|c|c}
  \hline
  Methods & \multicolumn{3}{c}{Test-R90}\\
  \cline{2-4}
  & PSNR & SSIM& FPS \\
  \hline
 Original & 16.15 (+0.00)  & 0.7407 (+0.0000)& --\\
 UDCP~\cite{UDCP} & 10.66 (-5.49)  & 0.4827 (-0.2580)& 0.1 \\
 WaterNet~\cite{water-net} & 16.11 (-0.04)  & 0.7872 (+0.0465) & 8.3\\
 Ucolor~\cite{Ucolor} & 21.04 (+4.89) & 0.8865 (+0.1458) & 1.7\\
 FUnIE-GAN~\cite{FUnIE-GAN} & 16.75 (+0.60) & 0.8160 (+0.0753) & 307 \\  
 UGAN~\cite{UGAN} & 21.03 (+4.88)& 0.8637 (+0.1230) & 190\\    
\hline
 \textbf{MTUR-Net} &  \textbf{22.60} (+6.45) & \textbf{0.9008} (+0.1601)& \textbf{30.3} \\
  \hline
\end{tabular}
}
\end{center}
\vspace{-15pt}
\end{table}

We conduct a full-reference evaluation using PSNR, SSIM, and FPS. Although the real-world environmental situation may differ from the reference image, the results of a fully-reference evaluation using the reference image can provide some feedback on the performance of different methods. A higher PSNR means that the result is less distorted, a higher SSIM means that the result is more similar to the reference image structure, and a higher FPS means that the processing process is more efficient. In Table \ref{tab:r90}, We can find that our method achieves the best PSNR and SSIM, while the FPS value is also ideal.

\begin{table}
\begin{center}
\caption{UIQM \cite{UIQM} Scores and UCIQE \cite{UCIQE} Scores of Different Methods on Test-C60} \label{tab:c60}\vspace{10pt}
\resizebox{\linewidth}{!}{
\begin{tabular}{c|c|c}
  \hline
  Methods & \multicolumn{2}{c}{Test-C60}\\
  \cline{2-3}
  & UIQM & UCIQE \\
  \hline
 Original & 1.898 (+0.000) & 0.5149 (+0.0000) \\
 UDCP~\cite{UDCP} & 1.156 (-0.742) & 0.5306 (+0.0157)\\
 WaterNet~\cite{water-net} & 2.405 (+0.507) & 0.5489 (+0.0340) \\
 Ucolor~\cite{Ucolor} & 2.450 (+0.552) & 0.5468 (+0.0319) \\
 FUnIE-GAN~\cite{FUnIE-GAN} & 2.704 (+0.806) & 0.5622 (+0.0473) \\
 UGAN~\cite{UGAN} & 3.114 (+1.216) & 0.5989 (+0.0840) \\
 \hline
 \textbf{MTUR-Net} &  \textbf{2.752} (+0.854) & \textbf{0.5868} (+0.0719) \\
  \hline
\end{tabular}
}
\end{center}
\vspace{-15pt}
\end{table}

Then we use UCIQE\cite{UCIQE} and UIQM\cite{UIQM} for a non-reference evaluation. In principle, the higher UCIQE score, the better balance of the standard deviation of the chroma, contrast of brightness, and average of saturation; for the higher UIQM score, the better the result is subjectively visually performed. In Table \ref{tab:c60}, our proposed model obtain one of the best scores in UCIQE and UIQM. However, when we visually compared the image with the first place, we found that there were many small squares on UGAN's image, but the score was still very high, indicating that this evaluation standard still needs to be improved.

\begin{figure}
    \centering
    \includegraphics[width=8cm]{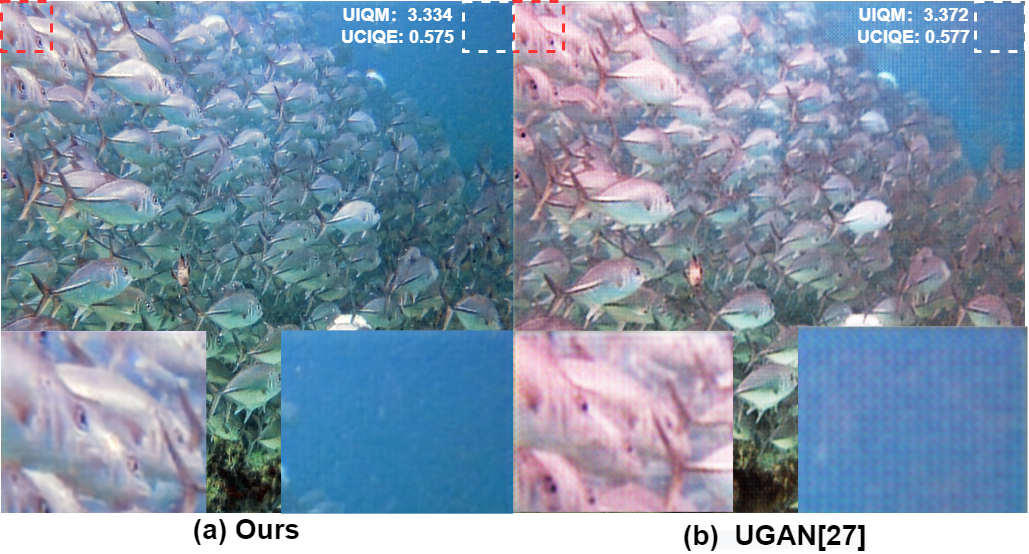}
    \caption{Test-C60 visual image comparison. Here we can see the difference between our image and the UGAN image. We don't have any obvious pixel cubes, and the contrast and color difference of objects are better.}
    \label{fig:survey}
\end{figure}

In order to further verify the effect of MTUR method, avoid the influence of subjective judgment of relevant experimenters on visualization results, and make our proposed method more convincing, besides quantitative  evaluation, we also conducted a series of research: We prepared 420 pictures expend from Test-C60 test set, each image corresponding seven different type(raw, MTUR, FUnIE-GAN, UGAN, Ucolor, UDCP and WaterNet) and then we vited 20 experimenters and asked them to compare the quality of the images in terms of chromatic aberration, visibility, clarity, etc., 
and select the best performance without knowing the corresponding experimental method of each image. After that we summarized in Table \ref{tab:fig}. As shown in the table, MTUR received the best rating in 42 of the 60 images in the TESTC-60 test set, and especially that MTUR generally obtains better recovery for details in a dark environment, combined with the image features.

\begin{table}[t]
\begin{center}

\caption{The generated image equality evaluation results of different methods on Test-C60.} \vspace{10pt}\label{tab:fig}
\includegraphics[width=8cm]{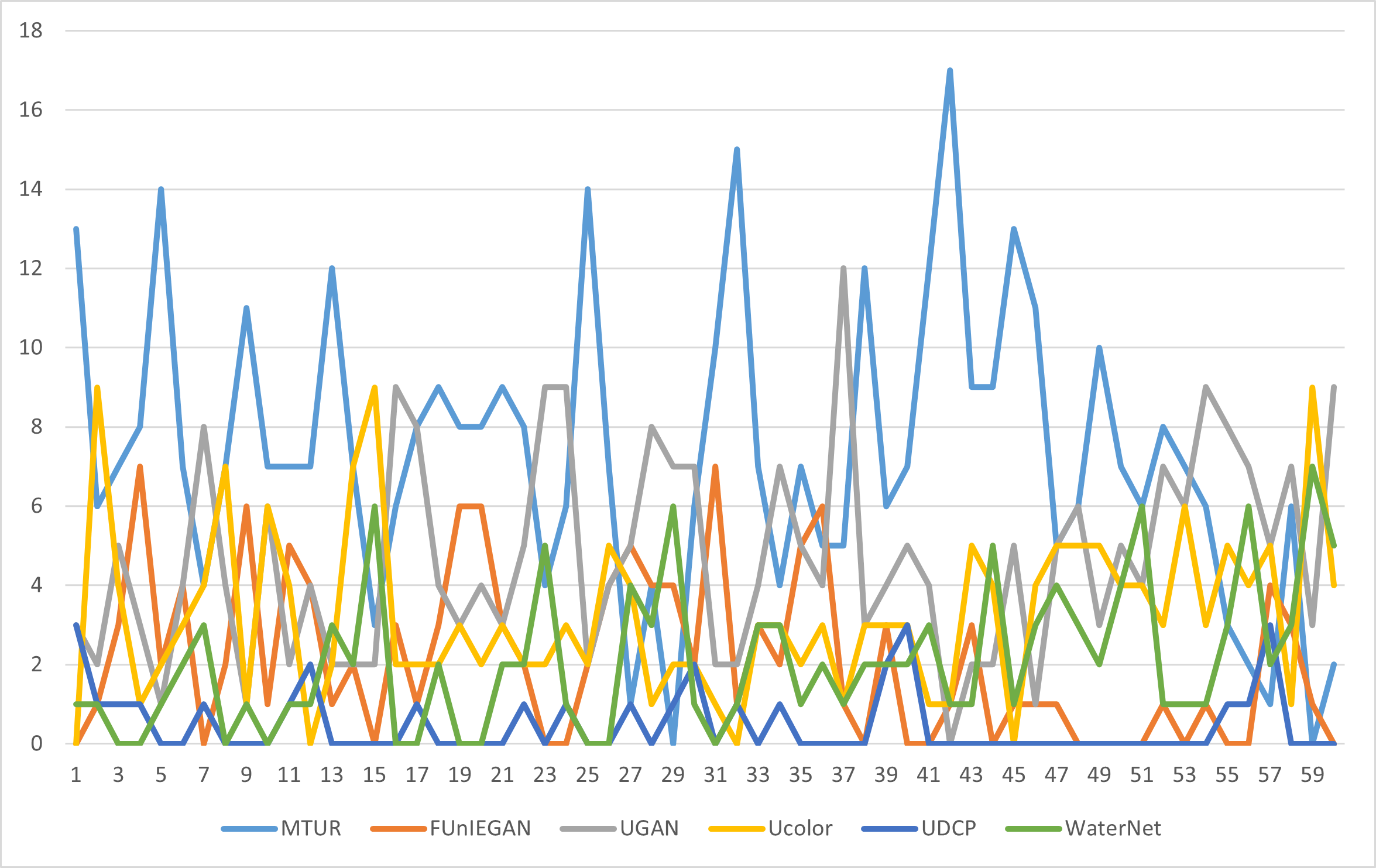}

\end{center}
\vspace{-15pt}
\end{table}

\subsection{Ablation Study}
We performed ablation experiments on test-R90 to verify the effectiveness of each part of the network. First, the first line’s basic network architecture is to remove the entire Medium Transmission Map module. So the network sustained the enhanced image directly based on the feature map generated from the dilated residual block (DRB) in underwater image enhancement subnet. The second line removes the skip connection between two subnetworks. Then, we did a comparative test to remove concatenation and only retain skip Connection. From the experimental results, we can find that without the final concatenation operation, the effect will be greatly reduced. Through these three experiments can prove that MT prediction subnetwork has a profound impact on image enhancement. After that, we try to reduce the convolution operation after concatenation, and we find that the effect also has an impact. In the last two ablation experiments, we tried to concatenate or add all DRB blocks together through skip connection to enhance the connection between shallow layer and deep layer network, and the results showed that the effect did not perform well.

\begin{table}[t]
\begin{center}

\caption{Component Analysis. The Basic Model is MTUR-Net without the MT-Guided Non-Local Module} \vspace{10pt}\label{tab:abl}
\resizebox{\linewidth}{!}{
\begin{tabular}{c|c|c}
  \hline
  
 Methods & PSNR & SSIM
  \\
  \hline
MTUR-Net &  22.60 (+0.00) & 0.9008 (+0.0000)\\
 \hline
Basic & 22.17 (-0.43) & 0.8893 (-0.0115) \\
w/o skip connection	& 22.15 (-0.45)	& 0.8897 (-0.0111) \\
w/o concat	& 21.63	(-0.97)& 0.8820 (-0.0188) \\
w/o conv2d after concat	& 22.23	(-0.37)& 0.8950 (-0.0058) \\
DRB blocks addition	& 21.75 (-0.85)	& 0.8890 (-0.0118)\\
DRB blocks concat & 21.92 (-0.67)& 0.8911 (-0.0097)\\
 \hline
\end{tabular}
}
\end{center}
\vspace{-15pt}
\end{table}

\section{Conclusion}\label{sec:conclusion}

In this paper, to solve the pain points existing in underwater image enhancement at this stage, we demonstrated the value of physical prior, in particular the medium transmission map, for restoring the real-world underwater images. By formulating a very simple network for learning both the prior and restoration results jointly, and encapsulating the knowledge interaction process across these two tasks at both feature and output levels, much better restoration features are learned thus guaranteeing much better results. Besides producing the best results on two real-world benchmarks, our model is also able to process the underwater images in a real-time speed, making it a potential framework to be deployed into intelligent underwater systems.

In the future, we will explore the upper-bound of benefits by the medium transmission map, and also continue the exploration of a more suited knowledge interaction design for better fusing the physical prior.

\bibliographystyle{IEEEbib}
\bibliography{icme2021template}

\begin{thebibliography}{10}

\bibitem{lee2012vision}
Donghwa Lee, Gonyop Kim, Donghoon Kim, Hyun Myung, and Hyun-Taek Choi,
\newblock ``Vision-based object detection and tracking for autonomous
  navigation of underwater robots,''
\newblock {\em Ocean Engineering}, vol. 48, pp. 59--68, 2012.

\bibitem{UDCP}
P.~Drews~Jr, E.~do~Nascimento, F.~Moraes, S.~Botelho, and M.~Campos,
\newblock ``Transmission estimation in underwater single images,''
\newblock pp. 825--830, 2013.

\bibitem{mohamed2011sensor}
Nader Mohamed, Imad Jawhar, Jameela Al-Jaroodi, and Liren Zhang,
\newblock ``Sensor {N}etwork {A}rchitectures for {M}onitoring {U}nderwater
  {P}ipelines,''
\newblock {\em Sensors}, vol. 11, no. 11, pp. 10738--10764, 2011.

\bibitem{Histogram}
Chong-Yi Li, Ji-Chang Guo, Run-Min Cong, Yan-Wei Pang, and Bo~Wang,
\newblock ``Underwater {I}mage {E}nhancement by {D}ehazing {W}ith {M}inimum
  {I}nformation {L}oss and {H}istogram {D}istribution {P}rior,''
\newblock {\em IEEE Transactions on Image Processing}, vol. 25, no. 12, pp.
  5664--5677, 2016.

\bibitem{water-net}
Chongyi Li, Chunle Guo, Wenqi Ren, Runmin Cong, Junhui Hou, Sam Kwong, and
  Dacheng Tao,
\newblock ``An {U}nderwater {I}mage {E}nhancement {B}enchmark {D}ataset and
  {B}eyond,''
\newblock {\em IEEE Transactions on Image Processing}, vol. 29, pp. 4376--4389,
  2020.

\bibitem{EUVP}
Md~Jahidul Islam, Youya Xia, and Junaed Sattar,
\newblock ``Fast {U}nderwater {I}mage {E}nhancement for {I}mproved {V}isual
  {P}erception,''
\newblock {\em IEEE Robotics and Automation Letters}, vol. 5, no. 2, pp.
  3227--3234, 2020.

\bibitem{SQUID}
D.~Berman, D.~Levy, S.~Avidan, and T.~Treibitz,
\newblock ``Underwater {S}ingle {I}mage {C}olor {R}estoration {U}sing
  {H}aze-{L}ines and a {N}ew {Q}uantitative {D}ataset,''
\newblock {\em IEEE Transactions on Pattern Analysis and Machine Intelligence},
  vol. 43, no. 08, pp. 2822--2837, aug 2021.

\bibitem{WaterGAN}
Jie Li, Katherine~A. Skinner, Ryan~M. Eustice, and Matthew Johnson-Roberson,
\newblock ``Watergan: {U}nsupervised {G}enerative {N}etwork to {E}nable
  {R}eal-{T}ime {C}olor {C}orrection of {M}onocular {U}nderwater {I}mages,''
\newblock {\em IEEE Robotics and Automation Letters}, vol. 3, no. 1, pp.
  387--394, 2018.

\bibitem{Ucolor}
Chongyi Li, Saeed Anwar, Junhui Hou, Runmin Cong, Chunle Guo, and Wenqi Ren,
\newblock ``Underwater {I}mage {E}nhancement via {M}edium
  {T}ransmission-{G}uided {M}ulti-{C}olor {S}pace {E}mbedding,''
\newblock {\em IEEE Transactions on Image Processing}, vol. 30, 2021.

\bibitem{8953954}
Xiaowei Hu, Chi-Wing Fu, Lei Zhu, and Pheng-Ann Heng,
\newblock ``Depth-{A}ttentional {F}eatures for {S}ingle-{I}mage {R}ain
  {R}emoval,''
\newblock in {\em 2019 IEEE/CVF Conference on Computer Vision and Pattern
  Recognition (CVPR)}, 2019, pp. 8014--8023.

\bibitem{UWCNN}
Chongyi Li, Saeed Anwar, and Fatih Porikli,
\newblock ``Underwater scene prior inspired deep underwater image and video
  enhancement,''
\newblock {\em Pattern Recognition}, vol. 98, pp. 107038, 2020.

\bibitem{UCycleGAN}
Chongyi Li, Jichang Guo, and Chunle Guo,
\newblock ``Emerging {F}rom {W}ater: {U}nderwater {I}mage {C}olor {C}orrection
  {B}ased on {W}eakly {S}upervised {C}olor {T}ransfer,''
\newblock {\em IEEE Signal Processing Letters}, vol. 25, no. 3, pp. 323--327,
  2018.

\bibitem{MDGAN}
Yecai Guo, Hanyu Li, and Peixian Zhuang,
\newblock ``Underwater {I}mage {E}nhancement {U}sing a {M}ultiscale {D}ense
  {G}enerative {A}dversarial {N}etwork,''
\newblock {\em IEEE Journal of Oceanic Engineering}, vol. 45, no. 3, pp.
  862--870, 2020.

\bibitem{CycleGAN}
Jun-Yan Zhu, Taesung Park, Phillip Isola, and Alexei~A. Efros,
\newblock ``Unpaired {I}mage-to-{I}mage {T}ranslation {U}sing
  {C}ycle-{C}onsistent {A}dversarial {N}etworks,''
\newblock in {\em 2017 IEEE International Conference on Computer Vision
  (ICCV)}, 2017, pp. 2242--2251.

\bibitem{he2010single}
Kaiming He, Jian Sun, and Xiaoou Tang,
\newblock ``Single {I}mage {H}aze {R}emoval {U}sing {D}ark {C}hannel {P}rior,''
\newblock {\em IEEE Transactions on Pattern Analysis and Machine Intelligence},
  vol. 33, no. 12, pp. 2341--2353, 2011.

\bibitem{huang2014visibility}
Shih-Chia Huang, Bo-Hao Chen, and Wei-Jheng Wang,
\newblock ``Visibility {R}estoration of {S}ingle {H}azy {I}mages {C}aptured in
  {R}eal-{W}orld {W}eather {C}onditions,''
\newblock {\em IEEE Transactions on Circuits and Systems for Video Technology},
  vol. 24, no. 10, pp. 1814--1824, 2014.

\bibitem{yang2011low}
Hung-Yu Yang, Pei-Yin Chen, Chien-Chuan Huang, Ya-Zhu Zhuang, and Yeu-Horng
  Shiau,
\newblock ``Low {C}omplexity {U}nderwater {I}mage {E}nhancement {B}ased on
  {D}ark {C}hannel {P}rior,''
\newblock in {\em 2011 Second International Conference on Innovations in
  Bio-inspired Computing and Applications}, 2011, pp. 17--20.

\bibitem{zhao2015deriving}
Xinwei Zhao, Tao Jin, and Song Qu,
\newblock ``Deriving inherent optical properties from background color and
  underwater image enhancement,''
\newblock {\em Ocean Engineering}, vol. 94, pp. 163--172, 2015.

\bibitem{fattal2008single}
Raanan Fattal,
\newblock ``Single {I}mage {d}ehazing,''
\newblock {\em ACM transactions on graphics (TOG)}, vol. 27, no. 3, pp. 1--9,
  2008.

\bibitem{8307410}
Yan-Tsung Peng, Keming Cao, and Pamela~C. Cosman,
\newblock ``Generalization of the {D}ark {C}hannel {P}rior for {S}ingle {I}mage
  {R}estoration,''
\newblock {\em IEEE Transactions on Image Processing}, vol. 27, no. 6, pp.
  2856--2868, 2018.

\bibitem{wu2018group}
Yuxin Wu and Kaiming He,
\newblock ``Group {N}ormalization,''
\newblock in {\em Proceedings of the European conference on computer vision
  (ECCV)}, 2018, pp. 3--19.

\bibitem{SELU}
G\"{u}nter Klambauer, Thomas Unterthiner, Andreas Mayr, and Sepp Hochreiter,
\newblock ``Self-{N}ormalizing {N}eural {N}etworks,''
\newblock Red Hook, NY, USA, 2017, NIPS'17, p. 972–981, Curran Associates
  Inc.

\bibitem{yu2017}
Fisher Yu, Vladlen Koltun, and Thomas Funkhouser,
\newblock ``Dilated {R}esidual {N}etworks,''
\newblock in {\em 2017 IEEE Conference on Computer Vision and Pattern
  Recognition (CVPR)}, 2017, pp. 636--644.

\bibitem{chen2017deeplab}
Liang-Chieh Chen, George Papandreou, Iasonas Kokkinos, Kevin Murphy, and
  Alan~L. Yuille,
\newblock ``Deeplab: Semantic {I}mage {S}egmentation with {D}eep
  {C}onvolutional {N}ets, {A}trous {C}onvolution, and {F}ully {C}onnected
  {C}rfs,''
\newblock {\em IEEE Transactions on Pattern Analysis and Machine Intelligence},
  vol. 40, no. 4, pp. 834--848, 2018.

\bibitem{wang2018understanding}
Panqu Wang, Pengfei Chen, Ye~Yuan, Ding Liu, Zehua Huang, Xiaodi Hou, and
  Garrison Cottrell,
\newblock ``Understanding {C}onvolution for {S}emantic {S}egmentation,''
\newblock in {\em 2018 IEEE Winter Conference on Applications of Computer
  Vision (WACV)}, 2018, pp. 1451--1460.

\bibitem{FUnIE-GAN}
Md~Jahidul Islam, Youya Xia, and Junaed Sattar,
\newblock ``Fast underwater image enhancement for improved visual perception,''
\newblock {\em IEEE Robotics and Automation Letters}, vol. 5, no. 2, pp.
  3227--3234, 2020.

\bibitem{UGAN}
Cameron Fabbri, Md~Jahidul Islam, and Junaed Sattar,
\newblock ``Enhancing underwater imagery using generative adversarial
  networks,''
\newblock in {\em 2018 IEEE International Conference on Robotics and Automation
  (ICRA)}. IEEE, 2018, pp. 7159--7165.

\bibitem{UCIQE}
Miao Yang and Arcot Sowmya,
\newblock ``An {U}nderwater {C}olor {I}mage {Q}uality {E}valuation metric,''
\newblock {\em IEEE Transactions on Image Processing}, vol. 24, no. 12, pp.
  6062--6071, 2015.

\bibitem{UIQM}
Karen Panetta, Chen Gao, and Sos Agaian,
\newblock ``Human-{V}isual-{S}ystem-{I}nspired {U}nderwater {I}mage {Q}uality
  {M}easures,''
\newblock {\em IEEE Journal of Oceanic Engineering}, vol. 41, no. 3, pp.
  541--551, 2016.

\end{thebibliography}

\end{document}